\documentclass[conference,10pt]{IEEEtran}

\IEEEoverridecommandlockouts

\usepackage{cite}
\usepackage{amsmath,amssymb,amsfonts}
\usepackage{graphicx}
\usepackage{textcomp}
\usepackage{booktabs}
\usepackage{cancel}
\usepackage{algorithm}
\usepackage{algpseudocode}
\usepackage{amsmath}
\usepackage{amssymb}
\usepackage{algorithm}
\usepackage{algpseudocode}
\usepackage{amsmath}
\usepackage{amssymb}
\usepackage{amsfonts}
\usepackage{microtype}
\usepackage{mathrsfs}%
\usepackage{multirow}
\usepackage{tabularx}
\usepackage{makecell}
\usepackage[normalem]{ulem} 
\usepackage{siunitx}
\usepackage[table]{xcolor}
\usepackage{float}
\usepackage{enumitem}
\usepackage{placeins}
\usepackage[hidelinks]{hyperref}
\setlist[itemize]{nosep,leftmargin=*}

\definecolor{bestgray}{gray}{0.80}
\definecolor{headergray}{gray}{0.92}
\definecolor{baselinegray}{gray}{0.92}
\definecolor{gold}{RGB}{255, 215, 0}
\definecolor{silver}{RGB}{192, 192, 192}
\definecolor{headerblue}{RGB}{30, 80, 162}
\definecolor{subheadergray}{RGB}{100, 120, 160}
\definecolor{modelrow}{RGB}{230, 238, 255}
\definecolor{denserow}{RGB}{220, 240, 220}
\definecolor{proprow}{RGB}{255, 245, 230}
\definecolor{reviewbrown}{RGB}{150, 75, 0}

\newcommand{\meanstd}[2]{\mbox{#1$\pm$#2}}

\def\BibTeX{{\rm B\kern-.05em{\sc i\kern-.025em b}\kern-.08em
    T\kern-.1667em\lower.7ex\hbox{E}\kern-.125emX}}

\makeatletter
\def\section{\@startsection{section}{1}{\z@}{0.9ex plus 0.4ex minus 0.2ex}{0.5ex plus 0.3ex minus 0ex}{\normalfont\normalsize\centering\scshape}}
\def\subsection{\@startsection{subsection}{2}{\z@}{0.9ex plus 0.4ex minus 0.2ex}{0.5ex plus 0.2ex minus 0ex}{\normalfont\normalsize\itshape}}
\def\subsubsection{\@startsection{subsubsection}{3}{\parindent}{0ex plus 0.1ex minus 0.1ex}{0ex}{\normalfont\normalsize\itshape}}
\makeatother
\begin{document}

\title{Multi-Objective Structured Pruning of LLMs for Latency and
  Model Size Optimization}

\author{\IEEEauthorblockN{Muhammad Junaid Ali\textsuperscript{1}, Smail Niar\textsuperscript{2}, El Ghazali Talbi\textsuperscript{3}}
\IEEEauthorblockA{\textsuperscript{1}Inria, University of Lille, Lille, France \\
\textsuperscript{2}LAMIH UMR CNRS 8201, UPHF, INSA Hauts-de-France, Valenciennes, France \\
\textsuperscript{3}University of Lille, CNRS, Inria, Centrale Lille, Lille, France \\
muhammad-junaid.ali@inria.fr, smail.niar@uphf.fr, el-ghazali.talbi@univ-lille.fr}
}

\maketitle
\thispagestyle{plain}
\pagestyle{plain}

\begin{abstract}
Large Language Models (LLMs) have achieved widespread adoption due to their strong reasoning and query-response capabilities. However, deploying them in embedded and edge computing environments remains challenging. Model pruning reduces complexity while preserving performance, but existing methods mainly optimize accuracy or parameter count, with latency treated as a post-pruning metric. Jointly pruning layers, attention heads, and Multi-Layer Perceptron (MLP) layers creates a large search space with varying accuracy-latency trade-offs. To address this challenge, we propose a hardware-aware multi-objective structured pruning framework for large language model compression. In the first stage, the framework identifies Pareto-optimal coarse pruning configurations by removing latency-costly sub-components while preserving performance. The second stage then applies Parallel Bayesian Optimization (PBO) to the Pareto-front solutions to allocate layer-wise sparsity under latency constraints, enabling broader exploration of latency-aware configurations. We benchmark the proposed method across multiple LLMs and sparsity ratios against existing pruning and sparsity allocation methods. Experimental results across four LLM families at 37.5$\%$ and 50$\%$ sparsity show that the proposed approach outperforms existing methods on downstream commonsense reasoning tasks on average, achieves lower perplexity on WikiText-2, C4, and FineWeb-Edu, and reduces 100-token inference latency to $0.716\,\mathrm{s}$ on an NVIDIA A100 GPU and $29.505\,\mathrm{s}$ on the evaluated Jetson Nano edge platform.

\end{abstract}

\section{Introduction}
Language models are not new, but transformer architectures have greatly accelerated the development of Large Language Models (LLMs). These models are designed to understand, generate, and predict human language, and they are now used in tasks such as machine translation, natural language generation, speech recognition, and optical character recognition. Some famous chatbots include ChatGPT, Claude, and Gemini. However, their large model size creates a major deployment challenge, especially for small edge devices or local environments with limited memory and strict latency requirements.
\par
Two primary approaches exist for obtaining small language models. The first is to train a compact architecture from scratch, but this often requires substantial data and computation. Alternatively, researchers can compress existing large models to preserve performance at a smaller scale. Existing studies have shown that LLMs contain substantial redundancy at different structural levels \cite{gromov2024unreasonable,men2025shortgpt,ma2023llm,ashkboos2024slicegpt}. In transformer-based LLMs, each block mainly consists of a Multi-Head Attention (MHA) module and a feed-forward Multi-Layer Perceptron (MLP) module. Since these modules differ in their impact on accuracy, parameter count, and latency, pruning can be applied at multiple granularities, from entire transformer blocks to individual attention heads or MLP neurons.

\par
Pruning approaches are typically divided into unstructured, semi-structured, and structured methods. Unstructured pruning removes individual weights and can yield high sparsity, but its irregular patterns often need specialized hardware or sparse kernels to improve speed. Structured pruning instead eliminates whole units, such as layers, attention heads, channels, or MLP neurons, creating dense smaller networks that run more efficiently on standard edge devices.

\par
Existing methods such as LLM-Pruner \cite{ma2023llm}, SliceGPT \cite{ashkboos2024slicegpt}, ShortGPT \cite{men2025shortgpt}, and BlockPruner \cite{zhong2025blockpruner} exploit structural redundancy to reduce model size while limiting quality loss. They use importance criteria or dimensionality reduction to select structures for removal. Staged methods such as 2SSP \cite{sandri20252ssp} and CFSP \cite{cfsp2025} further organize pruning into successive steps. However, these approaches leave two challenges insufficiently addressed. First, preserving accuracy and reducing parameter count do not by themselves ensure low latency on the target hardware, motivating explicit quality--latency optimization. Second, refining a single intermediate architecture limits exploration of how alternative coarse structures and layer-wise sparsity allocations affect the final trade-off.
\par
To address these limitations, we propose a two-stage framework that retains multiple coarse Pareto candidates for latency-aware sparsity allocation. Separating coarse architecture search from fine-grained refinement reduces the complexity of jointly searching layers, attention heads, and MLP dimensions.
\par
Therefore, Stage 1 performs coarse-grained Pareto search to remove MHA and MLP sub-blocks with lower performance contribution and higher latency or model-size cost. Stage 2 then refines each retained architecture through layer-wise sparsity allocation and sub-component pruning. This decomposition makes the search more efficient and stable, while the ablation studies evaluate whether the two-stage strategy achieves lower latency than using either stage independently.
\par

Table \ref{tab:comp_kd} places this two-stage framework in context by comparing its design with existing structured pruning methods. Among the compared methods, the proposed framework uniquely combines multi-objective search, layer-wise sparsity allocation, and edge evaluation.
\par

The contributions of this paper are as follows:
\begin{itemize}
\item We propose a two-stage, multi-objective structured pruning framework for LLM compression. The first stage identifies a Pareto set of layer-wise pruning configurations by jointly optimizing performance preservation and latency reduction. Instead of relying on a single Pareto candidate from the first stage, the second stage applies parallel Bayesian optimization over the returned Pareto-front solutions to identify a pruning configuration that maximizes performance retention while minimizing latency.

\item We propose a parallel Bayesian optimization strategy for layer-wise sparsity allocation, searching for pruning configurations that reduce inference latency while maintaining model performance. In contrast to prior methods that mainly emphasize accuracy preservation, our approach explicitly incorporates latency into the pruning search objective.

\item We benchmark the proposed method across multiple LLMs under different sparsity ratios. The results show that the proposed approach improves performance on downstream commonsense reasoning benchmarks under the evaluated sparsity budgets, achieves lower perplexity across multiple language-modeling evaluation datasets, and reduces inference latency while increasing throughput on both a high-performance GPU and an NVIDIA Jetson Nano edge device.
\end{itemize}
The remainder of this paper is structured as follows: Section II reviews related work, Section III presents the proposed methodology, Section IV describes the experimental settings, Section V reports the results and discussion, and Section VI concludes the paper.

\begin{table}[!t]
\centering
\footnotesize
\setlength{\tabcolsep}{1pt}
\begin{tabular}{lccccccc}
\hline
Method &
\makecell{Layer/\\block} &
\makecell{MHA} &
\makecell{MLP} &
\makecell{Two-\\stage} &
\makecell{Multi-obj.\\search} &
\makecell{Sparsity\\alloc.} &
\makecell{Edge\\eval.} \\
\hline
ShortGPT   \cite{men2025shortgpt} &
\checkmark & -- & -- & -- & -- & -- & -- \\
SliceGPT   \cite{ashkboos2024slicegpt} &
-- & \checkmark & \checkmark & -- & -- & -- & -- \\
BlockPruner   \cite{zhong2025blockpruner} &
\checkmark & \checkmark & \checkmark & -- & -- & -- & -- \\
2SSP   \cite{sandri20252ssp} &
-- & \checkmark & \checkmark & \checkmark & -- & -- & -- \\
CFSP   \cite{cfsp2025} &
\checkmark & \checkmark & \checkmark & \checkmark & -- & -- & -- \\
Proposed &
\checkmark & \checkmark & \checkmark & \checkmark & \checkmark & \checkmark & \checkmark \\
\hline
\end{tabular}
\caption{Structured pruning comparison. Layer/block, MHA, and MLP columns indicate pruning support.}
\label{tab:comp_kd}
\end{table}

\section{Related Work}
Early structured pruning approaches include LLM Pruner \cite{ma2023llm}, a task-agnostic method that removes non-essential coupled structures (dependency groups) based on gradient information, recovering performance through Low-Rank Adaptation (LoRA) fine-tuning. Similarly, Block Pruner performs fine-grained structured pruning within transformer blocks by pruning both MLP and MHA sub-components based on component-level importance \cite{zhong2025blockpruner}. Other methods focus on pruning model width and individual columns \cite{ma2023llm,sandri20252ssp}.
\par
SliceGPT uses Principal Component Analysis (PCA) to remove rows and columns across matrices \cite{ashkboos2024slicegpt}.
However, such approaches often require additional parameters or rigid structural adjustments to maintain compatibility. 2SSP \cite{sandri20252ssp} first prunes FFN neurons by output magnitude, then iteratively removes attention sub-modules, but this staged sequence may miss the global optimum. Similarly, BlockPruner \cite{zhong2025blockpruner} scores component importance to prune attention heads and MLPs. Unlike 2SSP, it applies block-level importance to both MLP and MHA pruning, but ignores sub-component-level pruning importance \cite{zhong2025blockpruner}.
\par
Recently, Coarse-to-Fine Structured Pruning (CFSP) \cite{cfsp2025} uses inter- and intra-block activation information to guide structured pruning. However, CFSP mainly relies on activation-based coarse-to-fine importance and does not explicitly optimize latency or model size as search objectives across multiple Pareto candidates. Existing layer-wise sparsity allocation methods mainly determine pruning ratios from layer importance, outlier statistics, reconstruction error, or theoretical error-propagation criteria \cite{yang2026lsa,li2024dsa,chen2025dlp,yin2024owl,huang2025determining}. Although some of these works report inference speedup after pruning, latency is generally treated as an evaluation metric rather than an explicit search objective. In contrast, our approach uses deployment-oriented objectives during pruning and refines multiple coarse Pareto candidates instead of relying on a single architecture.
\par
\begin{figure*}[!t]
    \raggedleft
    \includegraphics[width=1\linewidth]{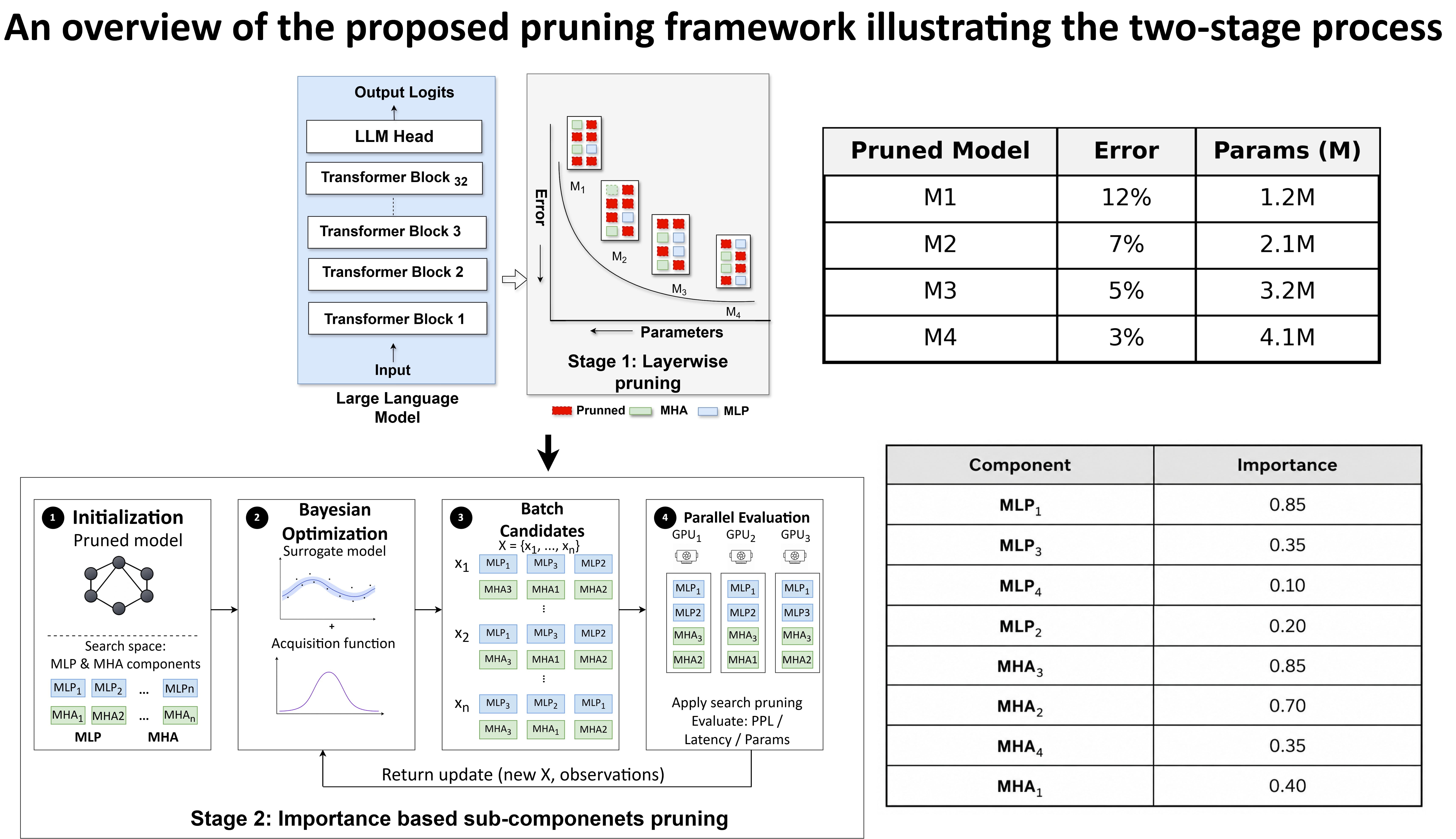}
    \caption{Overview of the proposed multi-objective structured pruning framework consisting of two different stages.}
    \label{fig:proposed}
\end{figure*}
\par      
\section{Proposed Methodology}
Our proposed methodology uses a global-to-local pruning framework that separates macro-level architecture search from fine-grained structural refinement. Figure \ref{fig:proposed} provides an overview of the methodology. The example illustrates how Stage 1 produces coarse Pareto candidates and Stage 2 refines them through latency-aware layer-wise allocation and sub-component pruning.
\par
Stage 1 prunes complete MHA and MLP sub-blocks to obtain coarse Pareto-optimal architecture candidates, while Stage 2 uses parallel Bayesian optimization to allocate the remaining sparsity budget and applies importance scoring to prune low-ranked sub-components.
\par
We next define the MHA and MLP notation used in the pruning formulation.
\begin{algorithm}[t]
\caption{Two-Stage Pruning with Bayesian Layer-Wise Allocation}
\label{alg:two_stage_pruning}
\begin{algorithmic}[1]

\State \textbf{Input:} Model $\mathcal{M}$, calibration data $\mathcal{D}$,
validation data $\mathcal{D}_{\mathrm{val}}$, sparsity budget $S$
\State \textbf{Output:} Best pruned model $M_{\mathrm{best}}$

\Statex \textbf{Stage 1: Multi-Objective Depth-Wise Pruning}

\State $\mathcal{Q} \gets \Call{InitPopulation}{}$

\While{stopping criterion is not met}
    \For{each candidate $c \in \mathcal{Q}$}
        \State $M_c \gets \Call{PruneSubBlocks}{\mathcal{M},c}$
        \State Evaluate $M_c$ using quality loss and latency
    \EndFor
    \State $\mathcal{P} \gets \Call{ParetoFront}{\mathcal{Q}}$
    \State $\mathcal{Q} \gets \Call{Evolve}{\mathcal{P}}$
\EndWhile

\State $\mathcal{C}_{\mathrm{sel}} \gets
\Call{SelectParetoCandidates}{\mathcal{P}}$

\Statex \textbf{Stage 2: Bayesian Allocation and Importance Benchmarking}

\State $\mathcal{A} \gets \Call{InitBayesianOptimizer}{S}$
\State $\mathcal{R} \gets \emptyset$
\Comment{Evaluated configurations}

\For{each coarse candidate $M_c \in \mathcal{C}_{\mathrm{sel}}$}
    \While{Bayesian optimization budget is not exhausted}
        \State Propose layer-wise allocations $\{a_1,\ldots,a_m\}$
        \For{each allocation $a_j$ \textbf{in parallel}}
            \For{each importance method $g$}
                \State Rank subcomponents using $g$ and $\mathcal{D}$
                \State $M_{c,j,g} \gets
                \Call{PruneByAllocation}{M_c,a_j,g}$
                \State Evaluate the perplexity of $M_{c,j,g}$
                on $\mathcal{D}_{\mathrm{val}}$ and measure its latency
                \State Add $M_{c,j,g}$ and its results to $\mathcal{R}$
            \EndFor
        \EndFor
        \State Update $\mathcal{A}$ using the observed perplexity
        and latency values
    \EndWhile
\EndFor

\State $M_{\mathrm{best}} \gets
\Call{SelectBestConfiguration}{\mathcal{R}}$
\Comment{Lowest perplexity and latency}

\State \Return $M_{\mathrm{best}}$

\end{algorithmic}
\end{algorithm}
\subsection{Multi-Head Attention}
Multi-Head Attention (MHA) is the sequence-context modeling module of a transformer block. It projects the input sequence into query, key, and value representations and allows each token representation to aggregate information from other token positions through multiple attention heads. Let $X \in \mathbb{R}^{S \times d_{\mathrm{model}}}$ denote the input hidden states, where $S$ is the sequence length and $d_{\mathrm{model}}$ is the hidden dimension. For head $h \in \{1, \dots, H\}$, the projections are defined as:
\begin{equation}
Q_h = XW_h^Q, \quad K_h = XW_h^K, \quad V_h = XW_h^V,
\end{equation}
where $W_h^Q, W_h^K \in \mathbb{R}^{d_{\mathrm{model}} \times d_k}$ and $W_h^V \in \mathbb{R}^{d_{\mathrm{model}} \times d_v}$ are head-specific projection matrices, and $d_k, d_v$ denote the query/key and value dimensions per head, respectively. The output of each head is computed as:
\begin{equation}
\mathrm{head}_h = \mathrm{softmax}\!\left(\frac{Q_h K_h^T}{\sqrt{d_k}}\right)V_h.
\end{equation}
The outputs of all $H$ heads are concatenated and passed through the output projection matrix $W^O \in \mathbb{R}^{(H \cdot d_v) \times d_{\mathrm{model}}}$:
\begin{equation}
\mathrm{MHA}(X) = \mathrm{Concat}(\mathrm{head}_1, \ldots, \mathrm{head}_H)W^O.
\end{equation}
In Stage 2, all importance estimators are applied under the same structured pruning unit. For MHA, the unit is an attention-head group. For layer $l$ and head $h$, we define this structurally coupled group as:
\begin{equation}
G^{\mathrm{MHA}}_{l,h} = \{W^Q_{l,h}, W^K_{l,h}, W^V_{l,h}, W^O_{l,\mathcal{I}_h,:}\},
\end{equation}
where $\mathcal{I}_h$ denotes the index set of the hidden dimensions associated with head $h$ in the concatenated tensor, and $W^O_{l,\mathcal{I}_h,:}$ denotes the corresponding incoming row slice of the output-projection matrix. The importance estimators used in Stage 2 assign a score to each group $G^{\mathrm{MHA}}_{l,h}$, and the lowest-scoring heads are removed according to the layer-wise pruning ratios proposed by Bayesian optimization. This group-wise definition allows weight-level estimators, such as Wanda-SP or gradient-based criteria, to be used consistently for structured head pruning by aggregating their scores over all parameters belonging to the same head.

\subsection{Feed-Forward Network}
The Feed-Forward Network (FFN), also referred to as the MLP module, applies token-wise nonlinear transformations after attention. Modern LLMs commonly use a gated MLP, such as SwiGLU, consisting of up, gate, and down projections. For an input hidden state $X \in \mathbb{R}^{S \times d_{\mathrm{model}}}$, the MLP computation is formulated as:
\begin{equation}
\mathrm{MLP}(X) = \left(\mathrm{SiLU}(XW_{\mathrm{Gate}}) \odot XW_{\mathrm{Up}}\right)W_{\mathrm{Down}},
\end{equation}
where $W_{\mathrm{Gate}}, W_{\mathrm{Up}} \in \mathbb{R}^{d_{\mathrm{model}} \times d_{\mathrm{ffn}}}$ expand the hidden representation to the intermediate dimension $d_{\mathrm{ffn}}$, $W_{\mathrm{Down}} \in \mathbb{R}^{d_{\mathrm{ffn}} \times d_{\mathrm{model}}}$ projects it back to the model dimension, and $\odot$ denotes element-wise multiplication. 

In Stage 2, MLP pruning removes structured intermediate-neuron groups. For layer $l$ and intermediate neuron $u \in \{1, \dots, d_{\mathrm{ffn}}\}$, the pruning group is defined as:
\begin{equation}
G^{\mathrm{MLP}}_{l,u} = \{W^{\mathrm{Gate}}_{l,:,u}, W^{\mathrm{Up}}_{l,:,u}, W^{\mathrm{Down}}_{l,u,:}\},
\end{equation}
where $W^{\mathrm{Gate}}_{l,:,u}$ and $W^{\mathrm{Up}}_{l,:,u}$ denote the $u$-th column vectors of the gate and up projections, and $W^{\mathrm{Down}}_{l,u,:}$ represents the $u$-th row vector of the down projection. Thus, pruning neuron $u$ consistently removes the coupled gate, up, and down projection components associated with that neuron. 

\subsection{Stage 1: Multi-Objective Layer-Wise Pruning}

Given an LLM $\mathcal{M}$ consisting of $\mathcal{L}$ layers, the objective is to identify the structural sub-blocks, specifically the Multi-Head Attention (MHA) and Feed-Forward Network (FFN) modules, to be pruned. Let $l \in \{1, 2, \dots, \mathcal{L}\}$ denote the layer index, and let $x_{l,\mathrm{MHA}}, x_{l,\mathrm{MLP}} \in \{0, 1\}$ be binary decision variables indicating whether the respective MHA and FFN modules are retained ($1$) or pruned ($0$) in layer $l$. The decision space $\Theta$ encompasses all possible pruning configurations across the entire model architecture. For each layer $l$, the decision space consists of two binary variables, yielding the search space:
\begin{equation}
\Theta = \prod_{l=1}^{\mathcal{L}} \{0,1\}^2 = \{0,1\}^{2\mathcal{L}}.
\end{equation}

The structural search is formulated as a multi-objective optimization problem aimed at finding the decision vector $x = \{ x_{1,\mathrm{MHA}}, x_{1,\mathrm{MLP}}, \ldots, x_{\mathcal{L},\mathrm{MHA}}, x_{\mathcal{L},\mathrm{MLP}} \} \in \Theta$ that simultaneously minimizes performance degradation and measured inference latency:
\begin{equation}
\min_{x \in \Theta} \Big( \mathcal{O}_1(x), \mathcal{O}_2(x) \Big),
\qquad p(x)\leq S,
\end{equation}
where $\mathcal{O}_{1}(x)$ denotes the performance loss between the original and pruned models, $\mathcal{O}_{2}(x)$ represents measured inference latency on the target hardware, and $p(x)$ is the removed-parameter fraction constrained by the target sparsity $S$.

To quantify the performance loss $\mathcal{O}_{1}(x)$, we measure the Kullback-Leibler (KL) divergence between the output distributions of the original model $\mathcal{M}_{\mathrm{original}}$ and the pruned model $\mathcal{M}_{\mathrm{pruned}}$. The KL divergence over the vocabulary token space $\mathcal{V}$ is formulated as:
\begin{equation}
D_{\mathrm{KL}}(P \parallel Q) = \sum_{v \in \mathcal{V}} P(v) \log \left(\frac{P(v)}{Q(v)}\right),
\end{equation}
where $P(v)$ and $Q(v)$ express the output probability distributions of $\mathcal{M}_{\mathrm{original}}$ and $\mathcal{M}_{\mathrm{pruned}}$, respectively, for a given context.

\begin{figure}[!htbp]
    \centering
    \includegraphics[width=\columnwidth]{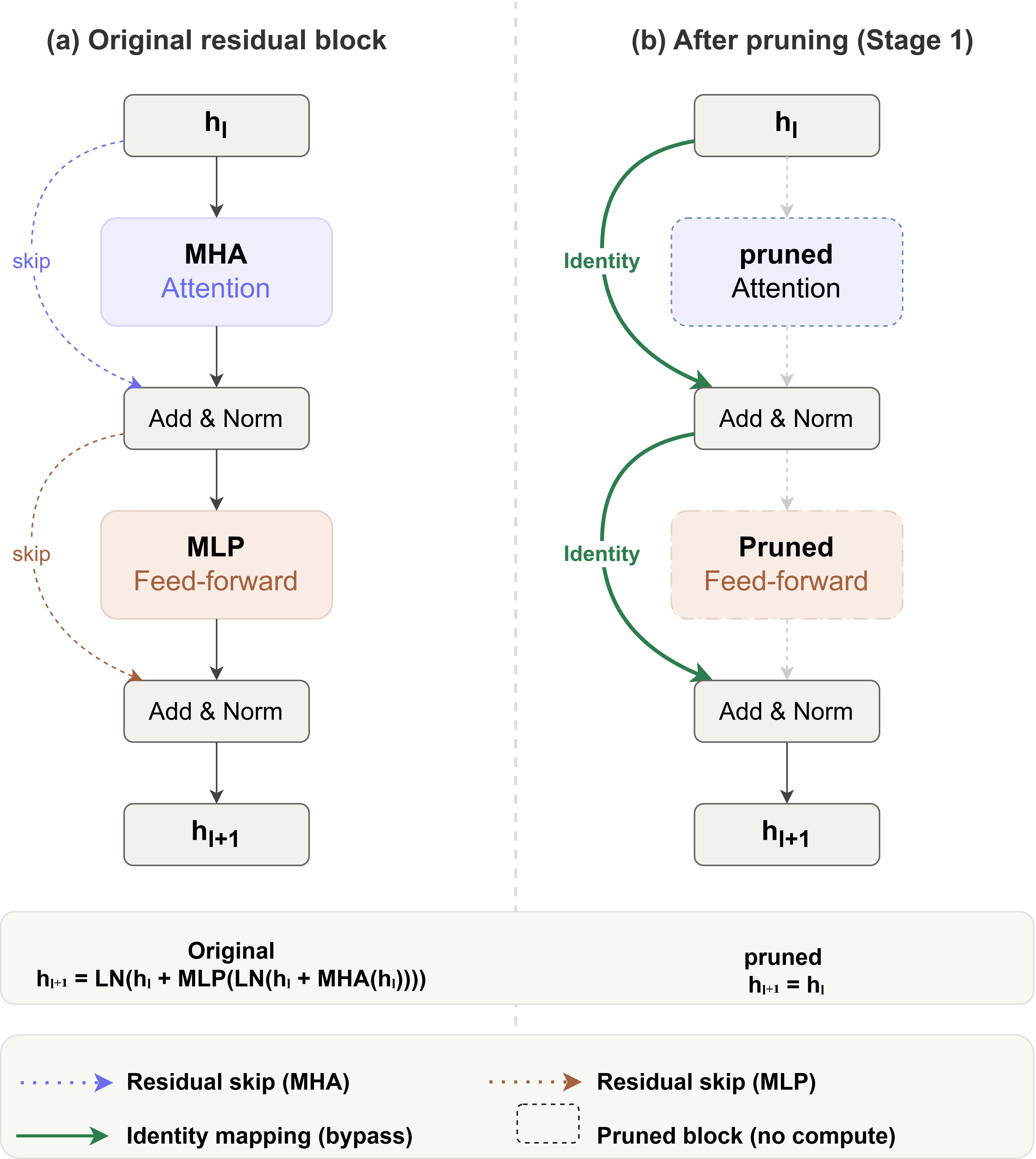}
    \caption{Block-level pruning in LLMs. (a) Standard transformer block with MHA and MLP. (b) Pruned block where computation is skipped and replaced with an identity connection.}
    \label{fig:identity}
\end{figure}

Layer pruning in Stage 1 is executed by replacing the forward pass of targeted sub-blocks with an identity mapping. By utilizing the identity mapping pathways within the residual blocks, the pruned layers are effectively bypassed, allowing the hidden states to propagate through the network via the skip connections without further computation, as illustrated in Figure \ref{fig:identity}.

\subsection{Stage 2: Parallel Bayesian Layer-Wise Allocation}
Stage 2 takes the Pareto-front candidates returned by Stage 1 and further optimizes the retained architectures. It searches over the remaining layer-wise pruning budgets of these candidates and evaluates the resulting models on WikiText-2. The feasible candidate with the lowest validation perplexity is selected, with measured latency used as the secondary criterion.

\par
The motivation behind using Bayesian optimization in Stage 2 is that the best layer-wise sparsity allocation is not known in advance. Fixed or uniform sparsity ratios may be suboptimal because different layers and sub-components contribute differently to accuracy, latency, and parameter reduction. Bayesian optimization guides the search toward promising allocations under latency and performance constraints, while the parallel setting evaluates multiple candidate allocations per iteration to reduce sequential evaluation bottlenecks.
\par
Let $a=\{a_l^{\mathrm{MHA}},a_l^{\mathrm{MLP}}\}_{l=1}^{\mathcal{L}}$ denote a candidate allocation vector, where $0\leq a_l^{\mathrm{MHA}},a_l^{\mathrm{MLP}}\leq 1$ are the pruning ratios assigned to MHA heads and MLP neurons in layer $l$. The allocation search  jointly reduces validation perplexity and latency:
\begin{equation}
\mathbf{f}(a)=
\left(
\mathrm{PPL}_{\mathrm{val}}\!\left(M_a\right),
T\!\left(M_a\right)
\right),
\quad
M_a=\mathrm{Prune}(M^{(1)},a,g).
\end{equation}
The feasible allocation set is defined by the remaining pruning budget:
\begin{equation}
\mathcal{A}_{\mathrm{feas}}=
\left\{a\in\mathcal{A}:\left|\mathcal{B}(a)-S_{\mathrm{rem}}\right|\leq\epsilon\right\},
\end{equation}
where candidate allocations are ranked according to their validation perplexity and
latency scores. Let
\[
\tau = (\tau_1,\ldots,\tau_{|\mathcal{A}_{\mathrm{feas}}|})
\]
denote the permutation of feasible-candidate indices obtained by sorting the
allocations in ascending lexicographic order of validation perplexity and
latency:
\begin{equation}
\tau
=
\operatorname{argsort}_{i \in
\{1,\ldots,|\mathcal{A}_{\mathrm{feas}}|\}}
\left(
\mathrm{PPL}_{\mathrm{val}}\!\left(M_{a_i}\right),
T\!\left(M_{a_i}\right)
\right).
\end{equation}
The optimal allocation is then selected as the highest-ranked feasible
candidate:
\begin{equation}
a^{*}=a_{\tau_1}.
\end{equation}
Thus, candidates with lower validation perplexity are preferred, with lower latency used to break ties. Here, $M^{(1)}$ denotes a Stage-1 Pareto candidate, $g$
denotes an importance strategy, $\mathrm{PPL}_{\mathrm{val}}(\cdot)$ denotes
validation perplexity, and $T(\cdot)$ denotes measured latency.
Furthermore, $S_{\mathrm{rem}}$ is the remaining pruning budget, while
$\epsilon$ accounts for rounding continuous pruning ratios to feasible
structured units, such as integer attention heads and MLP neurons. The budget
function $\mathcal{B}(a)$ maps the layer-wise pruning ratios to the corresponding
fraction of removed parameters. For each allocation, the importance strategy
ranks the subcomponents within the same Bayesian allocation framework. This
procedure preserves the multi-objective nature of the search while providing a
deterministic, deployment-oriented selection criterion.


\section{Experimental Settings}
All experiments were conducted on three NVIDIA A100 80GB GPUs using PyTorch 2.1, Hugging Face Transformers, lm-evaluation-harness, GPyTorch, and BoTorch. Evaluations were performed in a zero-shot setting on Massive Multitask Language Understanding (MMLU), HellaSwag (HS), AI2 Reasoning Challenge (ARC)-Easy/ARC-Challenge, Physical Interaction Question Answering (PIQA), and WinoGrande (WG). We report average accuracy across these benchmarks. For edge benchmarking, we used an NVIDIA Jetson Nano and report its measured latency separately from the A100 results.
\par
Table \ref{tab:experimental_settings} summarizes search settings. Baselines include ShortGPT, Sliding Window, BlockPruner, EvoPress, SliceGPT, 2SSP and CFSP. Table \ref{tab:results} reports CFSP results only for LLaMA-2-7B. Jetson Nano latency is measured after quantizing the pruned models to 4-bit precision using bitsandbytes. Table \ref{tab:latency_comparison} reports 10-run mean latency for 100-token sequences at 37.5\% sparsity. Ablations use 512-token sequences for computing the latency.

\begin{table}[!t]
\centering
\caption{Core search settings.}
\label{tab:experimental_settings}
\footnotesize
\setlength{\tabcolsep}{3pt}
\begin{tabular}{@{}p{0.30\columnwidth}p{\dimexpr0.70\columnwidth-6pt\relax}@{}}
\toprule
Parameter & Setting \\
\midrule
Stage 1 & NSGA-II (Pymoo) \cite{deb2002nsga2} \\
Population / budget & 40 individuals / 40 generations \\
Crossover / mutation & Simulated binary / mask mutation \\
Stage-1 calibration & 1,024 random WikiText-2 samples \\
Stage 2 & BoTorch/GPyTorch; \texttt{SingleTaskGP}, RBF kernel \\
Acquisition & Log expected improvement; 256 raw samples \\
BO budget & 20 warm-up + 20 BO trials per candidate \\
Target sparsity & 37.5\% and 50\%, searched independently \\
Constraints & Remaining budget $S-p_i$; layer retention \\
\bottomrule
\end{tabular}
\end{table}
\par
Across four models, we run each search independently for each target sparsity rate, 37.5$\%$ and 50$\%$. Stage 1 uses NSGA-II with a population of 40 for 40 generations, giving 1,600 evaluations per sparsity rate and 3,200 in total. It takes about 4 hours per sparsity rate. Stage 1 returns 8–10 candidates on average, so 18 architectures are evaluated downstream. Stage 2 uses 20 warm-up and 20 BO trials per candidate, giving 40 evaluations per candidate and 720 Stage-2 evaluations in total. It takes about 4 hours per sparsity rate. With two GPUs, this corresponds to about 14 GPU-hours per sparsity rate and 28 GPU-hours for both. Existing studies are mostly knowledge-driven and usually take 2–3 hours, while optimization-based methods such as EvoPress take about 48 hours. However, EvoPress targets unstructured sparsity and does not consider latency or layer-wise sparsity, which our optimization approach benefits from.
\par
We define the pruning ratio $S\in[0,1]$ as the target sparsity fraction corresponding to $100S\%$ pruning of the structured components or parameters considered in the search, following standard practices in existing studies \cite{men2025shortgpt, sandri20252ssp}. The distribution of this pruning budget between Stage 1 and Stage 2 is determined by the Stage-1 Pareto solutions, which may remove different fractions of the model.
\par
In Stage 1, pruning is applied up to the target budget $S$, generating multiple feasible coarse-pruned solutions. Each solution $i$ is characterized by its achieved sparsity fraction $p_i$, defined as:
\begin{equation}
    p_i = \frac{\text{Pruned parameters}_i}{\text{Total parameters}}
\end{equation}
Stage 2 then allocates the remaining pruning budget, $S_{\mathrm{rem}}=S-p_i$, for each respective solution using parallel Bayesian optimization. The optimized allocation specifies how much additional sub-component pruning should be applied within the retained layers to obtain a lower-latency configuration while satisfying the overall sparsity target.
\par
We evaluated the proposed method at 37.5\% and 50\% pruning rates across four LLM families to assess performance under different sparsity levels and model architectures. Comparisons with existing studies were conducted without LoRA recovery fine-tuning, highlighting the effectiveness of the pruning strategy itself. Since some baselines may rely on different implementations or recovery procedures, these results are interpreted as controlled pruning comparisons under the reported evaluation protocol rather than exhaustive baseline re-implementations.

\section{Results and Discussion}
\subsection*{Experimental Methodology}
We evaluate the proposed approach using two complementary criteria: accuracy on downstream tasks and perplexity on language-modeling datasets. We compare the approach with multiple structured-pruning baselines.

\par
Given that our proposed approach utilizes both depth-wise and component-wise pruning, we benchmark against the structural pruning baselines reported in Table \ref{tab:results}, including ShortGPT \cite{men2025shortgpt}, Sliding Window \cite{ding2025sliding}, Block Pruner \cite{zhong2025blockpruner}, EvoPress \cite{sieberlingevopress}, SliceGPT \cite{ashkboos2024slicegpt}, 2SSP \cite{sandri20252ssp} and CFSP \cite{cfsp2025}, a coarse-to-fine pruning approach. We compare against CFSP only on LLaMA-2-7B, as their method supports only Llama-based architectures.
\par
We evaluate two calibration settings: WikiText-2, reported as \textit{Proposed}, and a mixed set containing samples from each downstream task, reported as \textit{Proposed+Mix}. We conduct five independent runs for each proposed variant and report the mean and standard deviation; baseline values are included in the same table for comparison.
\par
For perplexity and downstream tasks, we perform experiments at sparsity rates of 37.5$\%$ and 50$\%$. Following recent LLM compression studies that use similar pruning budgets \cite{sandri20252ssp,ashkboos2024slicegpt}, this allows us to examine whether moderate and aggressive pruning preserve general reasoning performance and language-modeling quality as the pruning regime becomes more challenging.

\subsection*{Experimental Results}
\begin{table*}[!t]
\centering
\caption{Average zero-shot benchmark results for four LLMs at 37.5$\%$ and 50$\%$ sparsity levels. Gold and silver shading denote the best and second-best pruned results, respectively. Proposed results are the mean and standard deviation across five independent runs. Mix denotes mixed calibration.}
\label{tab:results}
\begingroup
\scriptsize
\setlength{\tabcolsep}{3pt}
\renewcommand{\arraystretch}{0.88}
\resizebox{0.94\textwidth}{!}{%
\begin{tabular}{@{}lrrrrrrrr@{}}
\toprule
Method & \multicolumn{2}{c}{Mistral-v0.3} & \multicolumn{2}{c}{LLaMA-2 7B} & \multicolumn{2}{c}{Qwen-2.5} & \multicolumn{2}{c}{Phi-3 14B} \\
\cmidrule(lr){2-3}\cmidrule(lr){4-5}\cmidrule(lr){6-7}\cmidrule(l){8-9}
& 37.5\% & 50\% & 37.5\% & 50\% & 37.5\% & 50\% & 37.5\% & 50\% \\
\midrule
\rowcolor{denserow}\textit{Dense} & 67.08 & 67.08 & 60.71 & 60.71 & 68.72 & 68.72 & 72.19 & 72.19 \\
ShortGPT \cite{men2025shortgpt}  & 38.13 & 36.83 & 42.63 & 36.89 & 41.28 & 32.33 & 48.34 & 34.83 \\
Sliding Window \cite{ding2025sliding} & 38.79 & 33.65 & 41.69 & 34.96 & 42.20 & 31.36 & 47.80 & 32.97 \\
Block Pruner \cite{zhong2025blockpruner} & 42.14 & 37.16 & 42.96 & 37.19 & 44.78 & 35.14 & 48.16 & 40.61 \\
EvoPress   \cite{sieberlingevopress} & 44.48 & 37.51 & 43.76 & 38.53 & 45.09 & 36.22 & 48.48 & 38.46 \\
SliceGPT \cite{ashkboos2024slicegpt} & 41.90 & 36.34 & 44.83 & 37.72 & 41.56 & 35.51 & 48.57 & 39.23 \\
2SSP \cite{sandri20252ssp}  & 45.73 & 39.27 & 47.60 & 37.76 & 46.37 & 41.47 & 49.10 & 43.17 \\
CFSP \cite{cfsp2025} & -- & -- & 49.62 & 36.94 & -- & -- & -- & -- \\
\rowcolor{proprow}Proposed & \cellcolor{silver!50}\underline{\meanstd{49.15}{0.24}} & \cellcolor{silver!50}\underline{\meanstd{39.69}{0.54}} & \cellcolor{silver!50}\underline{\meanstd{49.79}{0.65}} & \cellcolor{silver!50}\underline{\meanstd{38.65}{0.63}} & \cellcolor{silver!50}\underline{\meanstd{48.91}{0.74}} & \cellcolor{silver!50}\underline{\meanstd{43.39}{0.19}} & \cellcolor{silver!50}\underline{\meanstd{49.45}{0.35}} & \cellcolor{silver!50}\underline{\meanstd{44.64}{0.47}} \\
\rowcolor{proprow}Proposed+Mix & \cellcolor{gold!60}\textbf{\meanstd{51.92}{0.73}} & \cellcolor{gold!60}\textbf{\meanstd{46.03}{0.69}} & \cellcolor{gold!60}\textbf{\meanstd{52.99}{0.17}} & \cellcolor{gold!60}\textbf{\meanstd{40.39}{0.36}} & \cellcolor{gold!60}\textbf{\meanstd{51.84}{0.29}} & \cellcolor{gold!60}\textbf{\meanstd{45.34}{0.41}} & \cellcolor{gold!60}\textbf{\meanstd{50.09}{0.83}} & \cellcolor{gold!60}\textbf{\meanstd{45.96}{0.85}} \\
\bottomrule
\end{tabular}
}
\endgroup
\end{table*}
\begin{figure}[!htbp]
    \centering
    \includegraphics[width=1\columnwidth]{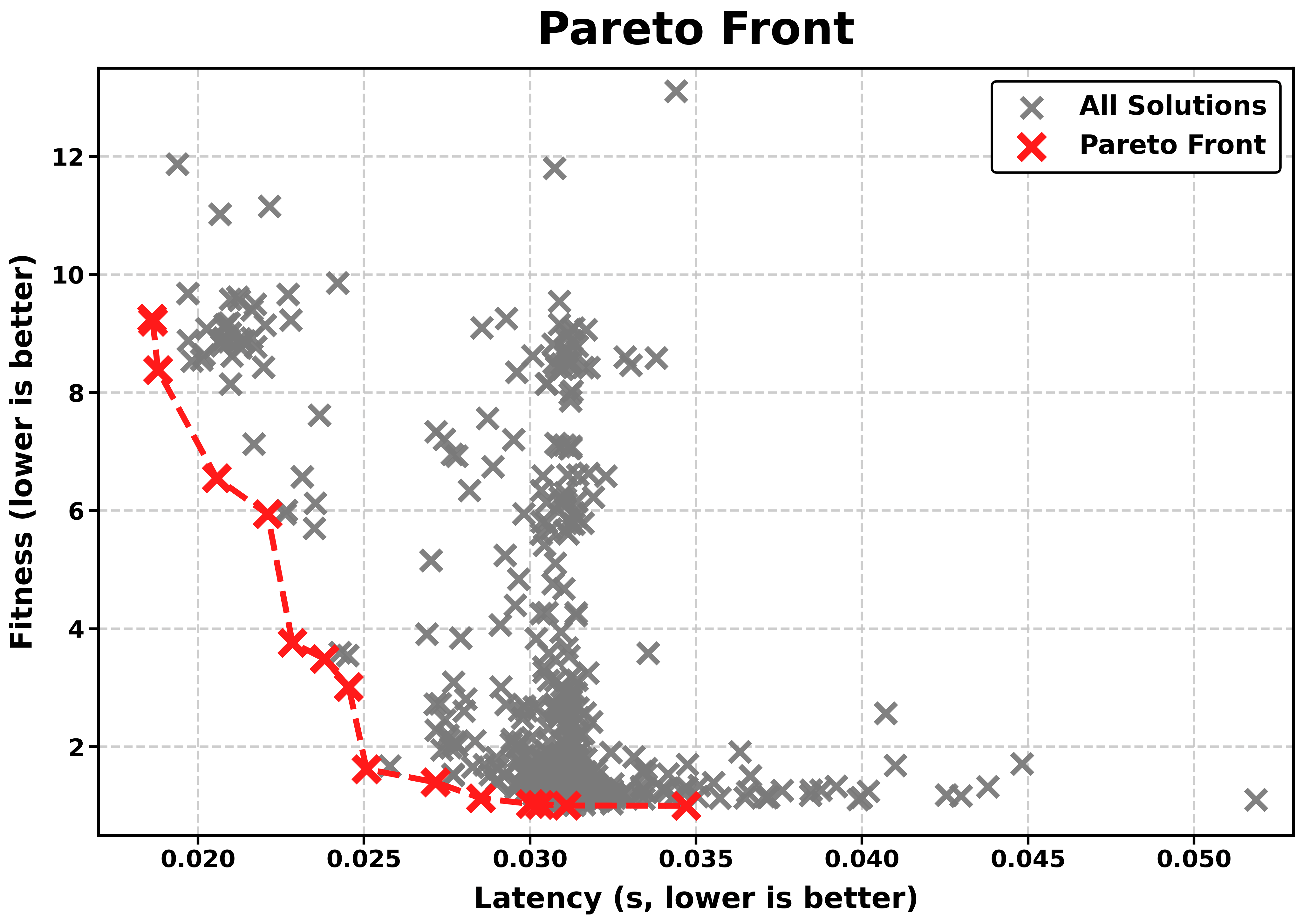}
    \caption{Visualization of the Pareto front and non-Pareto solutions at a sparsity rate of 37.5$\%$.}
    \label{fig:pareto_first}
\end{figure}

Figure \ref{fig:pareto_first} shows the visualization of solutions obtained after the multi-objective search on 37.5$\%$ pruning ratio, including individual solutions and Pareto front solutions. The accuracy on downstream tasks is reported for four popular models: LLaMA 2-7B \cite{touvron2023llama2}, Qwen 2.5 7B \cite{ahmed2025qwen}, Phi-3 14B \cite{abdin2024phi} and Mistral-v0.3-7B   \cite{jiang2023mistral}. The detailed results are provided in Table \ref{tab:results}.
\par
The results at 37.5$\%$ and 50$\%$ sparsity show that the proposed approach achieves the strongest overall average downstream accuracy across all four model families. In particular, it consistently preserves the aggregate downstream performance under structured pruning, indicating that the compressed models retain a robust accuracy profile across the evaluated setting. At 50$\%$ sparsity, competing approaches degrade more sharply, whereas the proposed approach still retains the highest average accuracy across all models.
\par
\begin{figure}[!htbp]
    \centering
    \includegraphics[width=1\columnwidth]{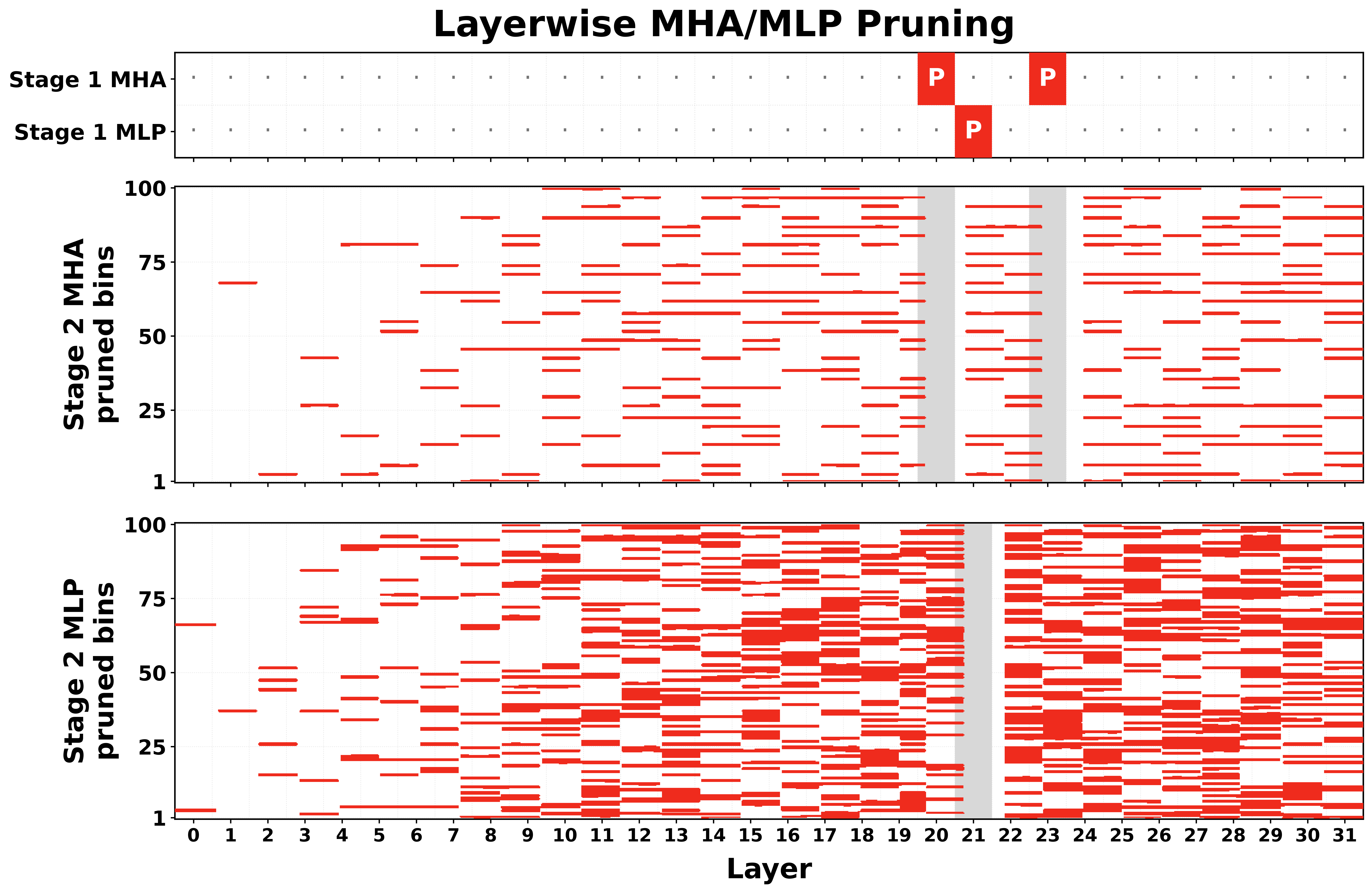}
    \caption{Layer-wise MHA and MLP pruning patterns under the two-stage strategy at 37.5$\%$ sparsity.}
    \label{fig:pruning_patterns}
\end{figure}

At 50$\%$ sparsity, the method remains especially strong in terms of average downstream accuracy across the evaluated model families. The results of mixed calibration sets are reported as  \textit{Proposed+Mix}. Figure \ref{fig:pruning_patterns} visualizes the layer-wise MHA and MLP pruning patterns for the 37.5\% sparsity setting. The pruning distribution is non-uniform across layers and components, demonstrating that BO does not apply a fixed pruning ratio universally. Instead, it removes more MHA or MLP capacity from layers that are less critical to the selected objectives, while preserving components that are highly sensitive to average accuracy drops. This behavior explains why the proposed method can reduce latency and active model size without causing the sharp downstream performance degradation typically observed in uniform or single-stage pruning strategies. 
\par
The results show that the proposed approach is not dependent on a single calibration source and performs well in both mixed calibration and WikiText-2.
\par
In addition to zero-shot accuracy, Table \ref{tab:latency_comparison} reports end-to-end latency on 100-token sequences averaged over 10 runs on an NVIDIA A100 and the evaluated NVIDIA Jetson edge platform. These values refer to the full 100-token generation protocol; comparison with ablation timings requires matched batch size, prompt length, precision, and prefill/decode timing conventions. The proposed approach achieves the lowest latency on both platforms ($0.716\,\mathrm{s}$ on the A100 and $29.505\,\mathrm{s}$ on the Jetson Nano), confirming that the multi-objective search and parallel BO search improve deployment efficiency together with accuracy.
\par
\begin{table}[!ht]
\centering
\caption{Comparison of latency between the proposed approach and
existing studies. Results are shown in seconds, computed on
100-token text sequences over 10 iterations on an Nvidia A100
GPU and the evaluated Nvidia Jetson edge platform.}
\label{tab:latency_comparison}
\footnotesize
\setlength{\tabcolsep}{4pt}
\renewcommand{\arraystretch}{1.08}
\begin{tabularx}{\columnwidth}{@{}Xrr@{}}
\toprule
\rowcolor{headergray}
Method & A100 (s) & Jetson Nano (s) \\
\midrule
Sliding Window \cite{ding2025sliding} & 0.784 & 32.690 \\
2SSP \cite{sandri20252ssp} & 0.854 & 40.530 \\
BlockPruner \cite{zhong2025blockpruner} & 0.825 & 47.565 \\
EvoPress \cite{sieberlingevopress} & 0.891 & 44.625 \\
ShortGPT \cite{men2025shortgpt} & 0.792 & 49.875 \\
CFSP \cite{cfsp2025} & 0.753 & 31.325 \\
\rowcolor{baselinegray}
Proposed & \cellcolor{bestgray}\textbf{0.716} & \cellcolor{bestgray}\textbf{29.505} \\
\bottomrule
\end{tabularx}
\end{table}

Although the proposed approach maintains an overall balance across all evaluated metrics, certain performance gaps arise because Stage-1 prioritizes global latency and efficiency, while Stage-2 relies on task-agnostic importance scores from a general calibration set rather than optimizing for individual benchmarks. Consequently, it selectively prunes components that offer minimal performance contributions while incurring high computational costs. Despite these localized trade-offs, the proposed approach achieves the highest average performance alongside the lowest latency across the tested hardware.
\par
\begin{figure*}[!t]
    \centering
    \includegraphics[width=\textwidth]{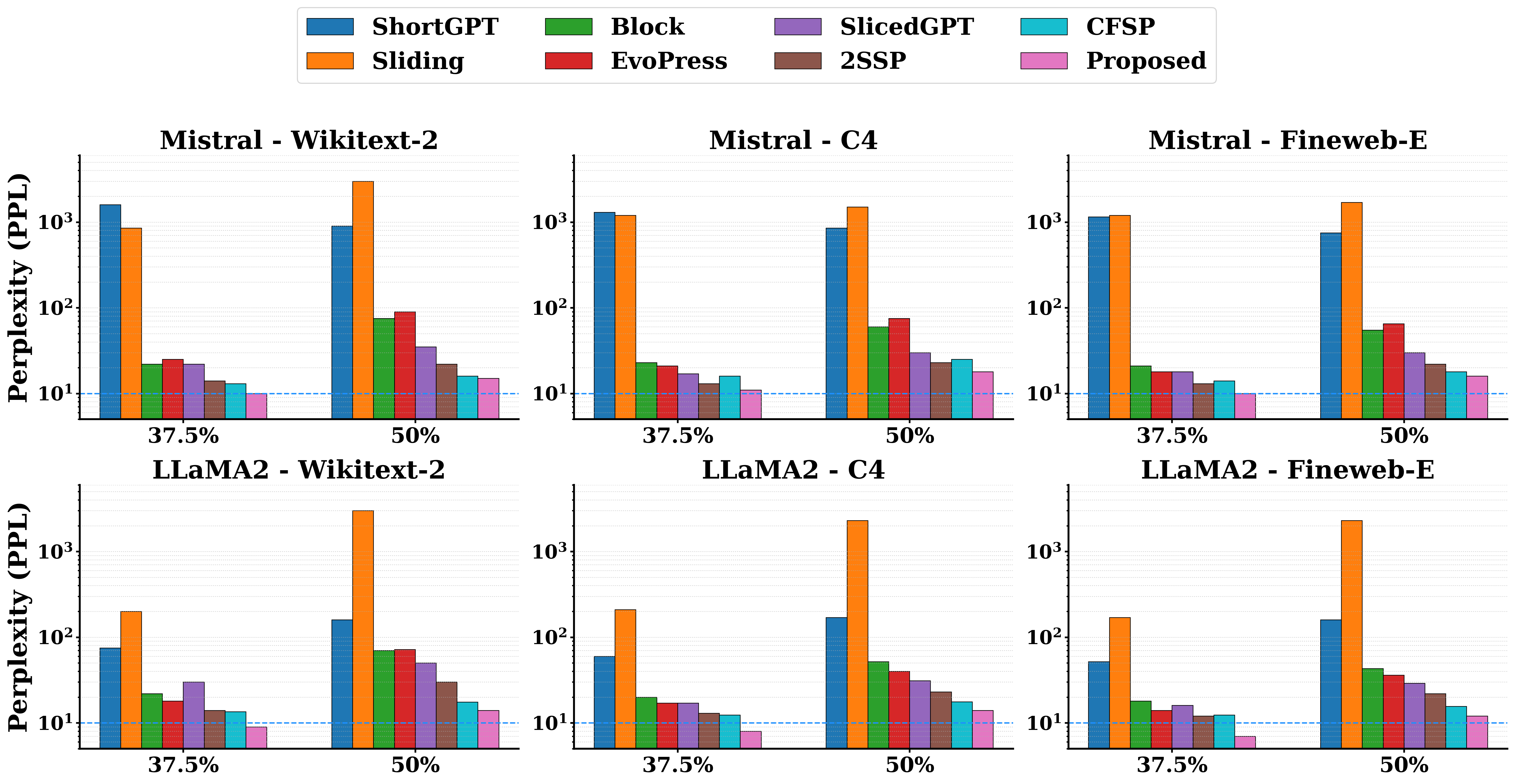}
    \caption{Perplexity (PPL) comparison across Mistral and LLaMA2 on WikiText-2, C4, and FineWeb-Edu.}
    \label{fig:ppl}
\end{figure*}

In addition to benchmarking downstream tasks, we computed perplexity (PPL) scores on three benchmarks: WikiText-2, FineWeb-Edu, and C4. The bar chart in Figure \ref{fig:ppl} compares these scores at two sparsity rates (37.5$\%$ and 50$\%$) for two LLMs, LLaMA-2-7B and Mistral-v0.3. The sub-figures show that the proposed approach generally achieves lower perplexity across the evaluated benchmarks and models, confirming its effectiveness beyond downstream-task accuracy. These results also complement the downstream accuracy and latency tables, where multiple runs are reported to reduce the influence of run-to-run variation and provide a more reliable comparison across pruning configurations.

\par
\begin{figure}[!htbp]
    \centering
    \includegraphics[width=0.8\columnwidth]{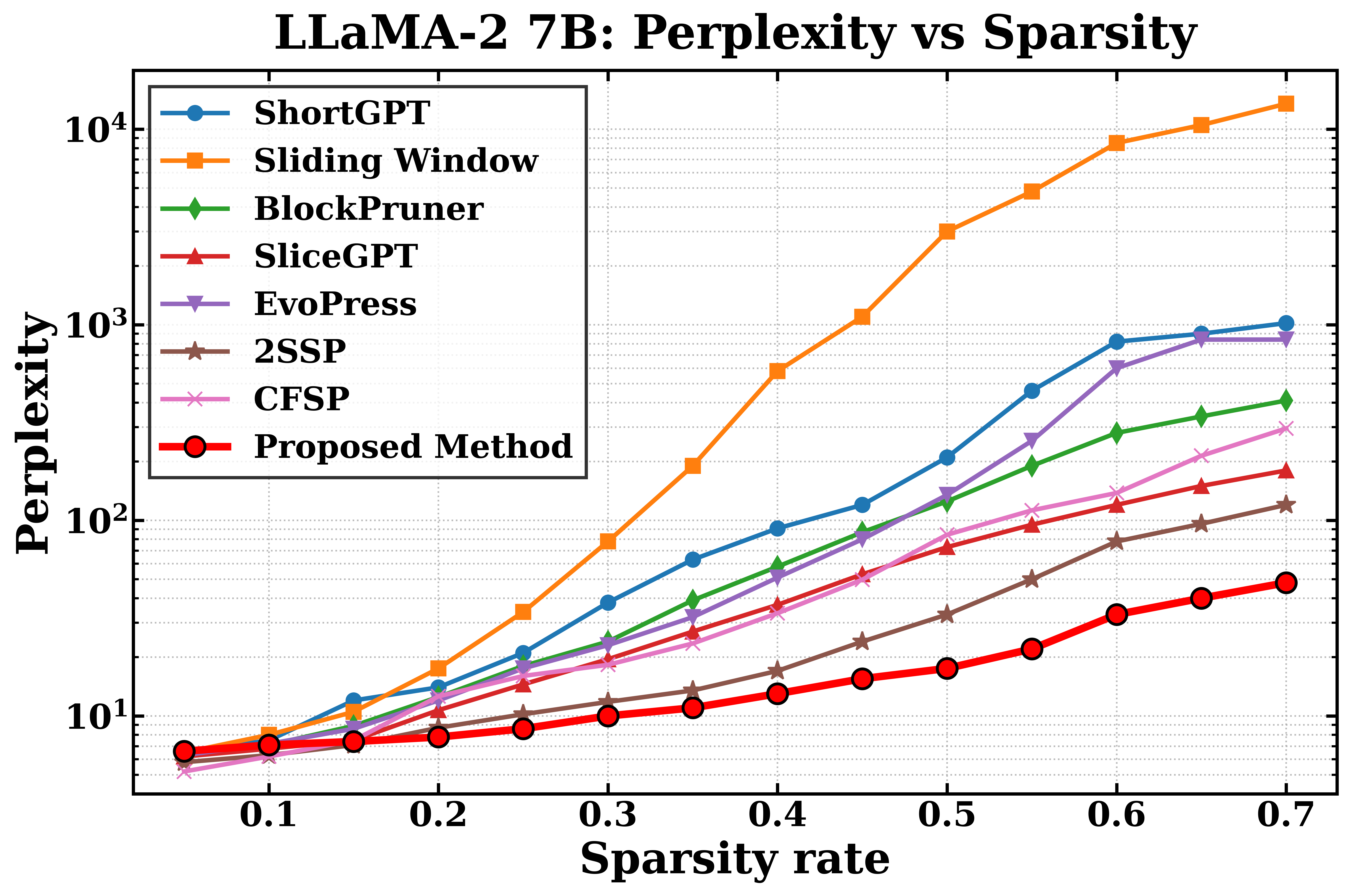}
    \caption{Comparison of the perplexity values in log scale on different sparsity rates of the proposed approach with existing studies }
    \label{fig:ppl_vs_sparsity}
\end{figure}

To study the impact of the sparsity rate ($S$) on performance, we calculate the perplexity scores on WikiText-2 at different sparsity rates after applying Stage 1 and Stage 2 and compare them with baseline studies, as shown in Figure \ref{fig:ppl_vs_sparsity}. The results show that varying $S$ significantly impacts performance; while smaller sparsity rates result in only a minor increase in perplexity, a substantial increase is observed beyond 40$\%$ sparsity, with the rise becoming more pronounced between 50$\%$ and 70$\%$.
\par
This is due to the removal of essential weights that encode the model's fundamental linguistic patterns, which cannot be recovered once the redundancy threshold is surpassed. The baseline methods did not show increased perplexity until 40$\%$, but after 40$\%$ they exhibit a significant increase; however, the proposed approach maintains a lower increase in perplexity compared to existing studies even after 40$\%$. This demonstrates that the proposed approach remains robust even as the sparsity rate increases.
\par

\subsection{Critical Difference Analysis}
\begin{figure}[H]
    \centering
    \includegraphics[width=\columnwidth]{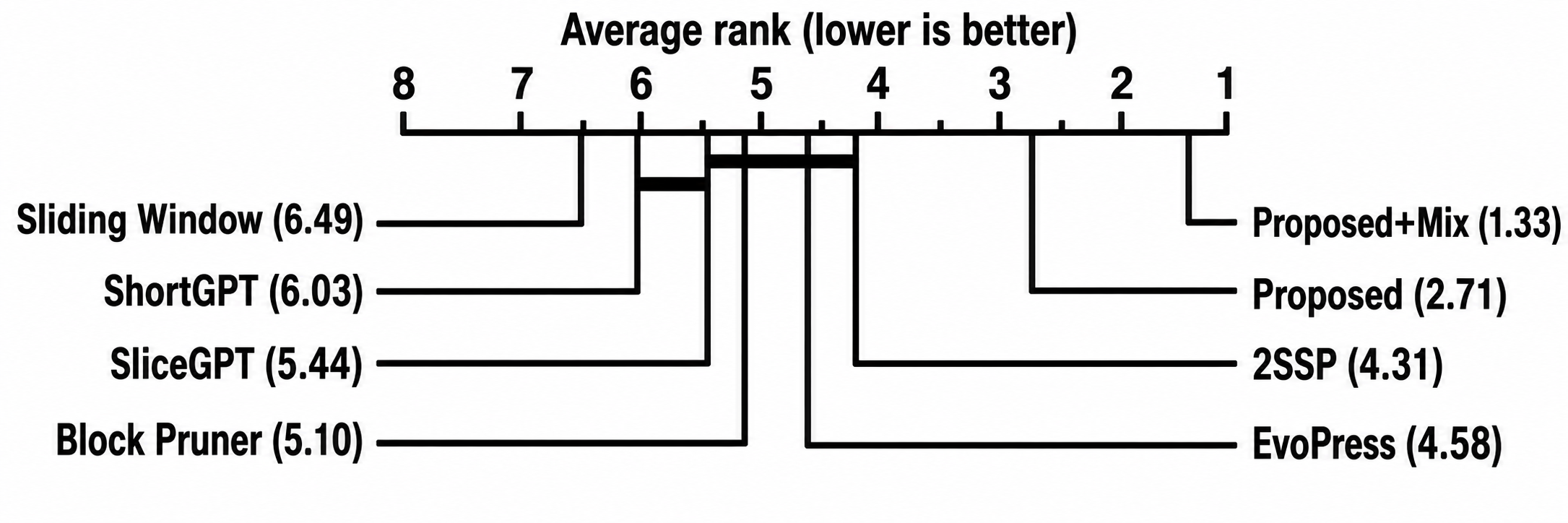}
    \caption{Critical difference diagram showing the pairwise statistical comparison of structured pruning methods with the proposed approach.}
    \label{fig:cd_diagram_llm_pruning}
\end{figure}

To assess the statistical significance of the differences between the proposed approach and existing structured pruning methods, we performed a rank-based Critical Difference analysis, excluding CFSP since it is reported only for LLaMA-2 7B. First, the Friedman test was applied to examine whether the compared methods differ significantly. After rejecting the null hypothesis, we performed a post-hoc Wilcoxon signed-rank test with Holm's correction at $\alpha = 0.05$. The resulting average ranks are visualized using a Critical Difference (CD) diagram, as shown in Figure \ref{fig:cd_diagram_llm_pruning}. Figure \ref{fig:cd_diagram_llm_pruning} shows that the proposed approach using the mixed calibration set achieves the best average rank, followed by the proposed approach using WikiText-2 calibration. The better rank achieved with mixed calibration can be attributed to improved pruning stability, as the mixed calibration set exposes the search process to more diverse activation patterns during the search.

\subsection{Ablation studies}
We conducted ablation studies to isolate the main design choices of the proposed framework. We performed ablation studies on LLaMA-2-7B. The ablation studies are conducted at 37.5\% sparsity and the proxy-correlation analysis additionally compares 37.5\% and 50\% sparsity to evaluate how the proxy behavior changes under more aggressive pruning. The analysis focuses on five primary questions: (i) whether the proxy metrics used during the search correlate with downstream accuracy (ii) how the calibration set affects downstream performance and latency (iii) whether the two-stage design is more effective than using either stage alone (iv) which objective combination provides the best accuracy–efficiency trade-off and (v) how different stage-2 sparsity allocation and importance estimation strategies affect the pruned model's performance.

\subsubsection{Proxy Metrics and Downstream Accuracy}
To validate our Stage 1 multi-objective search, we analyze the correlation between the resulting Pareto fronts (optimized for KL divergence and perplexity) and final downstream accuracy. We compute these proxy metrics for all candidate pruned models and compare them against their zero-shot performance. The motivation behind this analysis is to validate these proxies, as our search procedure relies on computationally inexpensive evaluations to filter candidates before selecting them for expensive downstream validation.
\begin{table}[!htbp]
\centering
\caption{Correlation and squared-correlation analysis with downstream average accuracy.}
\label{tab:correlation_downstream_accuracy}
\scriptsize
\setlength{\tabcolsep}{2.5pt}
\renewcommand{\arraystretch}{1.08}
\begin{tabular}{llrrrr}
\hline
Sparsity & Pred. & $\boldsymbol{r}$ & $\boldsymbol{r^2}$ & $\boldsymbol{\rho}$ & $\boldsymbol{\rho^2}$ \\
\hline
0.375 & KL div. & -0.967 & 0.935 & -0.794 & 0.630 \\
0.375 & PPL     & -0.358 & 0.128 & -0.006 & 0.000 \\
0.5   & KL div. &  0.417 & 0.174 &  0.381 & 0.145 \\
0.5   & PPL     & -0.828 & 0.685 & -0.786 & 0.617 \\
\hline
\end{tabular}
\end{table}

\begin{figure}[htp]
    \centering
    \includegraphics[
        width=\columnwidth,
        height=0.16\textheight
    ]{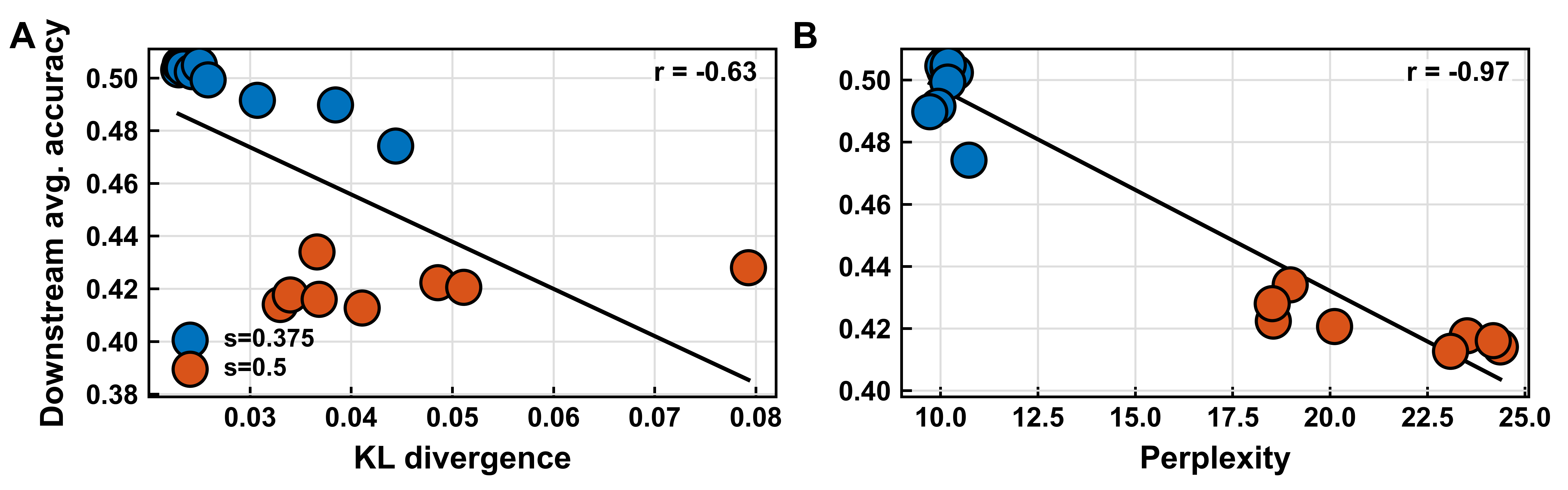}
    \caption{Correlation between evaluation proxies and downstream accuracy across sparsity levels.}
    \label{fig:proxy_correlation}
\end{figure}

As shown in Figure \ref{fig:proxy_correlation} and Table \ref{tab:correlation_downstream_accuracy}, KL divergence is a stronger proxy for downstream accuracy at moderate sparsity, whereas perplexity becomes more reliable under aggressive pruning. At 37.5$\%$ sparsity, KL divergence is strongly correlated with downstream accuracy ($r=-0.967$), but this relationship weakens at 50$\%$ sparsity, where perplexity shows a stronger association. This indicates that KL divergence is better suited to conservative pruning, while perplexity is more effective when pruning substantially affects language-modeling performance.


\subsubsection{Calibration-Set Size}

\begin{figure}[!b]
    \centering
    \includegraphics[width=1\columnwidth]{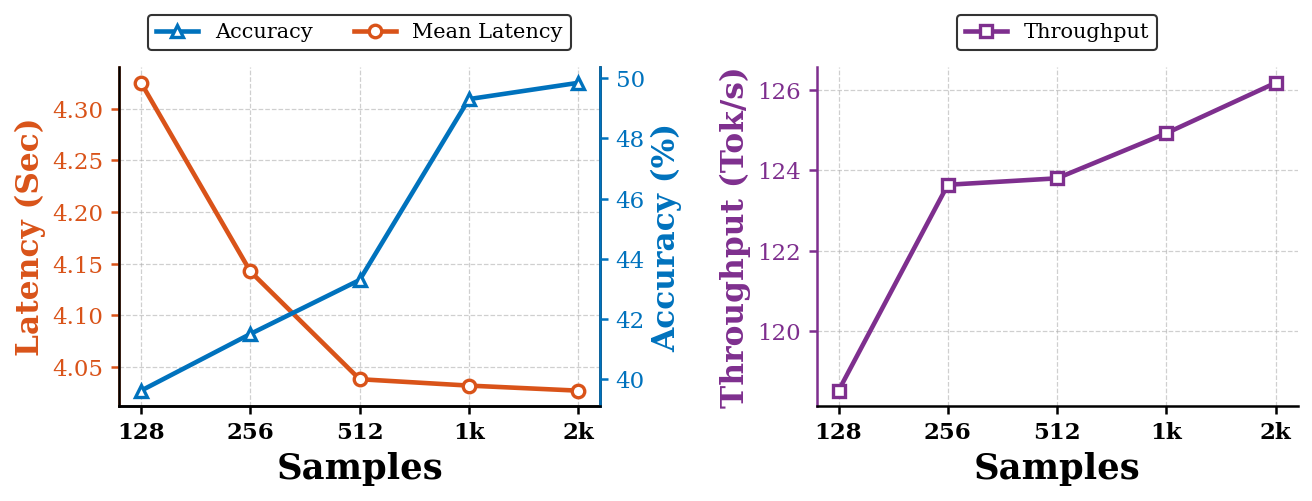}
    \caption{Effect of calibration sample size on latency, throughput, and downstream accuracy. }
    \label{fig:calibration_samples_tradeoff}
\end{figure}

The calibration size controls the quality of the activation statistics used for pruning. Figure \ref{fig:calibration_samples_tradeoff} summarizes the relationship between calibration size, downstream performance, latency, and throughput. It shows that increasing the calibration set from 128 to 2048 samples consistently improves deployment metrics: mean latency decreases from approximately $4.33\,\mathrm{s}$ to approximately $4.03\,\mathrm{s}$, while throughput increases from approximately 118.5 to approximately 126.2 tokens/s. Downstream accuracy follows the same trend, rising from approximately 39.7$\%$ at 128 samples to approximately 49.8$\%$ at 2048 samples. 
\par
The largest accuracy gain occurs between 512 and 1024 samples, suggesting that smaller calibration sets are insufficient to capture task-relevant behavior. However, the latency curve begins to saturate after 512 samples, showing that efficiency gains plateau earlier than accuracy gains. This result supports using sufficiently large calibration sets while highlighting the practical trade-off between calibration cost and marginal performance improvement. Given the diminishing returns between 1024 and 2048 samples, we selected a calibration size of 1024 for our experiments to avoid unnecessary computational run-time.

\subsubsection{Stage-Wise Combination Ablation}
\begin{figure}[htp]
    \centering
    \includegraphics[width=\columnwidth]{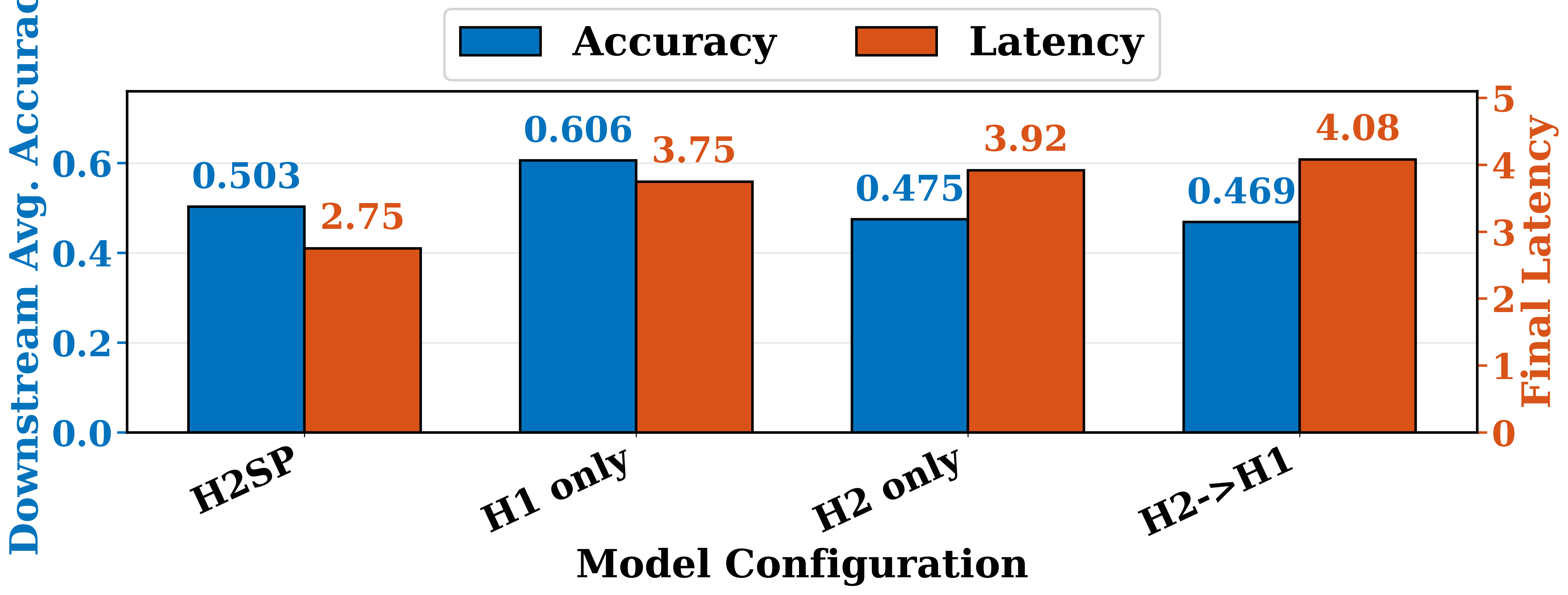}
    \caption{Downstream accuracy and latency for stage-wise pruning configurations.}
    \label{fig:component_scenarios}
\end{figure}

We evaluate whether a two-stage pruning approach is essential or if single-stage pruning is equally effective. To demonstrate this we have performed multiple experiments with alternative pipeline decompositions: depth-only, width-only, sequential, reversed, and combined configurations as shown in Figure \ref{fig:component_scenarios}.
The results show that depth pruning removes blocks while balancing performance preservation and efficiency, whereas width pruning alone offers fine control but misses block-level redundancy.  Combining the two allows the method to first identify globally efficient candidates and then refine them through component-level pruning to balance latency and accuracy.  This shows the importance of two-stage design. We also evaluate a reversed stage-wise setting, where width-wise pruning (Stage 2) is applied first, followed by depth pruning (Stage 1). The target sparsity is divided equally between the two stages, with 18.75$\%$ applied in each stage. In this setting, the depth-pruning candidates are ranked using both objectives, and the best candidate is selected. As shown in Figure \ref{fig:component_scenarios}, this reversed order gives the lowest downstream accuracy ($0.469$) and the highest latency ($4.08\,\mathrm{s}$), indicating that applying width-wise pruning first removes useful sub-components too early and yields a weaker accuracy-latency trade-off than the proposed H2SP order.

\subsubsection{Objective-Combination Analysis}
Our multi-objective formulation consists of two main objectives: one for model performance and one for efficiency. We evaluate two performance criteria (KL-divergence and perplexity) and two efficiency criteria (latency and number of parameters). The results for the different objective combinations are given in Table \ref{tab:obj_combo_ieee_onecol_no_thr}.
\begin{table}[!htbp]
\caption{Objective-combination comparison. Lower inference time and PPL are better; higher accuracy is better.}
\label{tab:obj_combo_ieee_onecol_no_thr}
\centering
\scriptsize
\setlength{\tabcolsep}{2.5pt}
\renewcommand{\arraystretch}{1.0}
\begin{tabular}{l l r r r}
\toprule
1st Objective & 2nd Objective &
Inference (s)$\downarrow$ &
Accuracy$\uparrow$ &
PPL$\downarrow$ \\
\midrule
Perplexity    & Params  & 7.95 & 0.5821 & 8.2943 \\
KL-Divergence & Params  & 5.86 & \cellcolor{bestgray}\textbf{0.5930} & 8.1393 \\
Perplexity    & Latency & 6.60 & 0.5732 & 9.2695 \\
KL-Divergence & Latency &
\cellcolor{bestgray}\textbf{4.86} &
0.5647 &
\cellcolor{bestgray}\textbf{7.48} \\
\bottomrule
\end{tabular}
\end{table}
\par
With respect to the first objective, KL-Divergence generally outperforms perplexity in efficiency and language-modeling performance, while the downstream-accuracy comparison depends on the second objective. When the number of parameters is used as the second objective, KL-Divergence yields lower inference time ($5.86\,\mathrm{s}$ vs. $7.95\,\mathrm{s}$), higher mean downstream accuracy (0.5930 vs. 0.5821, averaged over tasks), and lower average perplexity (8.14 vs. 8.29), computed over WikiText-2, C4, and FineWeb-Edu.
\par
When latency is used as the second objective, the optimization prioritizes runtime efficiency and perplexity over downstream accuracy while still maintaining a balanced trade-off between the objectives. In this setting, KL/Latency achieves the lowest inference time ($4.86\,\mathrm{s}$) and the lowest perplexity (7.49), at the cost of a reduced mean downstream accuracy of 0.565. Overall, these results reveal a clear accuracy--latency trade-off: KL/Params is preferable when accuracy preservation is prioritized, whereas KL/Latency is preferable under stringent latency constraints, offering an approximately 17\% reduction in inference time at the cost of a 0.028 absolute (2.8 percentage-point) decrease in mean downstream accuracy.

\subsubsection{Stage-2 Importance Estimator Ablation}
We perform a comparative study of different importance estimation strategies in Stage 2 while keeping Stage 1 unchanged. Table \ref{tab:stage2_pruning_metrics} compares Stage-2 pruning methods in terms of average downstream accuracy, average perplexity over WikiText-2, C4, and FineWeb-Edu, and latency.

\begin{table}[!htbp]
\centering
\caption{Comparison of Stage-2 importance estimators.}
\label{tab:stage2_pruning_metrics}
\begin{tabular}{lccc}
\toprule
Method & Downstream Avg. $\uparrow$ & Avg. PPL $\downarrow$ & Latency (s) $\downarrow$ \\
\midrule
Importance \cite{minitron2024}      & 0.4634 & 29.9688 & 3.7887 \\
Wanda-SP \cite{wanda}               & \textbf{0.4927} & \textbf{21.6250} & \textbf{3.7604} \\
Taylor \cite{ma2023llm}             & 0.4821 & 23.6250 & 3.7659 \\
Gradient \cite{huang2026gradpruner} & 0.4811 & 23.5312 & 3.7610 \\
WActiGrad \cite{chitty2024wactigrad}& 0.4413 & 26.3438 & 3.9623 \\
\bottomrule
\end{tabular}
\end{table}
Table \ref{tab:stage2_pruning_metrics} shows that Wanda-SP achieves the best result across all three reported metrics, with the highest downstream average accuracy ($0.4927$), the lowest average PPL ($21.6250$), and the lowest latency ($3.7604\,\mathrm{s}$). This indicates that the activation-aware salience metric of Wanda-SP \cite{wanda} preserves structures that are useful not only for downstream prediction but also for language-modeling perplexity and inference speed in this setting. Taylor and Gradient pruning provide the next-best downstream accuracy and perplexity, with Gradient achieving slightly lower PPL and latency than Taylor, while Taylor gives slightly higher downstream accuracy. In contrast, WActiGrad \cite{chitty2024wactigrad} obtains the lowest downstream accuracy ($0.4413$) and the highest latency ($3.9623\,\mathrm{s}$), making it the least favorable estimator in this ablation. Overall, these results support using Wanda-SP as the Stage 2 importance estimator for the evaluated configuration.

\subsubsection{Stage-2 Sparsity Allocation Methods Ablation}

\begin{table}[!htbp]
\centering
\caption{Stage-2 allocation-method comparison. Lower PPL and latency are better; higher throughput and downstream average are better.}
\label{tab:stage2_allocation_metrics}
\resizebox{\columnwidth}{!}{%
\begin{tabular}{lcccc}
\toprule
Method & Avg. PPL $\downarrow$ & Latency (s) $\downarrow$ & Throughput (tokens/s) $\uparrow$ & Downstream Avg. $\uparrow$ \\
\midrule
Uniform & 15.8694 & 3.8216 & 133.98 & 0.4572 \\
LSA \cite{yang2026lsa} & 13.8296 & 3.8948 & 131.60 & 0.4038 \\
DSA \cite{li2024dsa} & 12.0280 & 3.8965 & 131.47 & 0.4142 \\
DLP \cite{chen2025dlp} & 10.3632 & 3.8641 & 132.53 & 0.4898 \\
ER \cite{huang2025determining} & 10.4251 & 3.9467 & 129.87 & 0.4929 \\
OWL \cite{yin2024owl} & 10.3745 & 3.8689 & 132.35 & 0.4898 \\
ER+ \cite{huang2025determining} & 10.3632 & 3.8885 & 131.67 & 0.4898 \\
Proposed & \textbf{9.6309} & \textbf{3.7871} & \textbf{135.21} & \textbf{0.5142} \\
\bottomrule
\end{tabular}
}
\end{table}

Table \ref{tab:stage2_allocation_metrics} compares the Stage 2 sparsity-allocation strategies against the proposed allocation strategy. The proposed method achieves the best value for every reported metric, with the lowest average PPL ($9.6309$), lowest latency ($3.7871\,\mathrm{s}$), highest throughput ($135.21$ tokens/s), and highest downstream average accuracy ($0.5142$). Among the baselines, ER \cite{huang2025determining} gives the strongest downstream accuracy ($0.4929$) but has the highest latency and lowest throughput, while DLP \cite{chen2025dlp}, OWL \cite{yin2024owl}, and ER+ provide similar downstream accuracy ($0.4898$) with lower PPL than Uniform, LSA \cite{yang2026lsa}, and DSA \cite{li2024dsa}. These results indicate that the proposed parallel allocation search improves the accuracy-efficiency trade-off rather than optimizing only a single metric.

\begin{figure}[!htbp]
    \centering
    \includegraphics[width=\columnwidth]{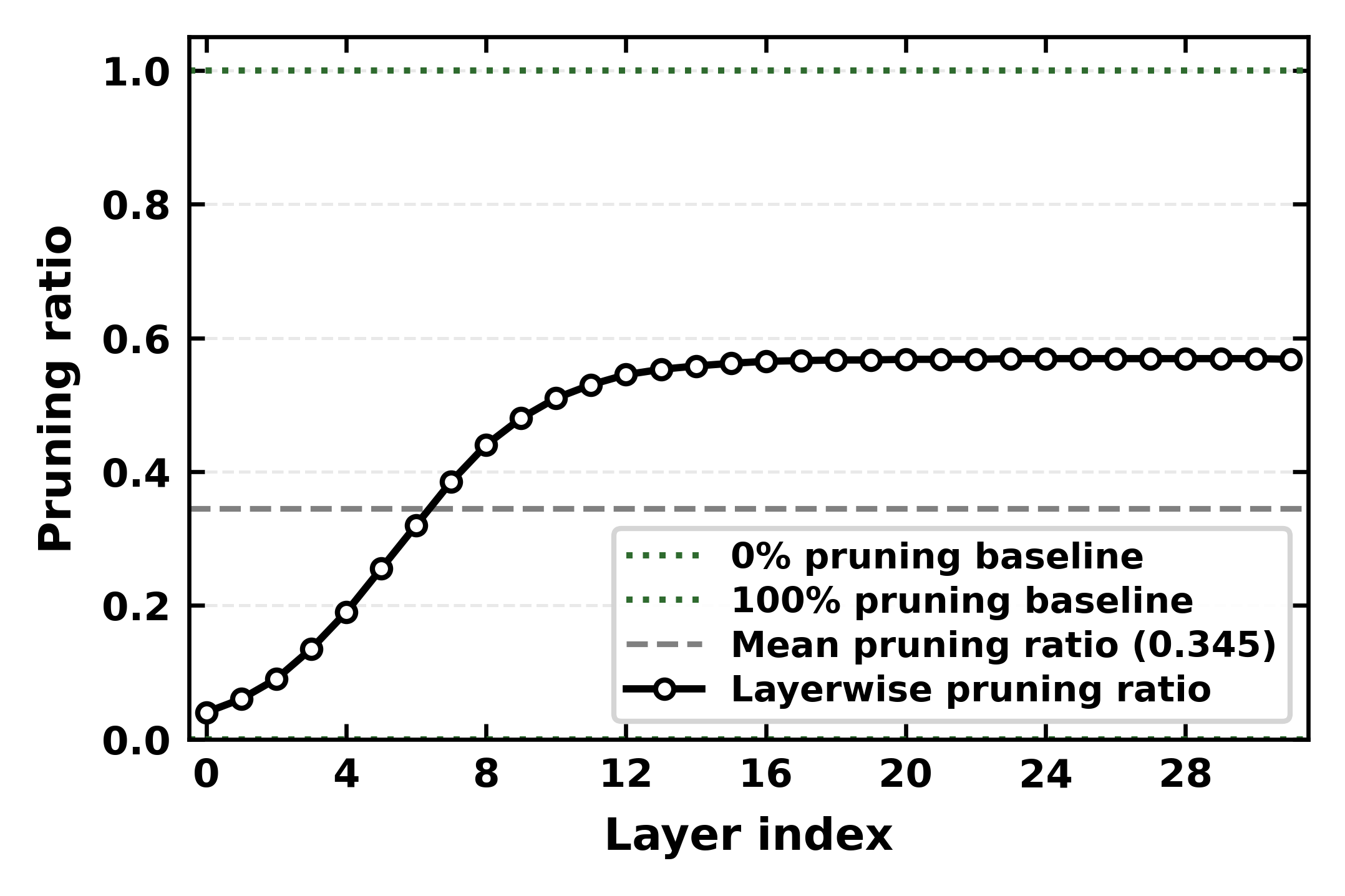}
    \caption{Layer-wise pruning ratios for the selected LLaMA-2-7B candidate.}
    \label{fig:layerwise_patterns}
\end{figure}

Figure \ref{fig:layerwise_patterns} shows that the selected LLaMA-2-7B candidate uses non-uniform layer-wise pruning ratios.

\section{Conclusion}
We presented a hardware-aware two-stage structured pruning framework that improves average commonsense-reasoning performance and provides competitive or lower perplexity across four LLM families and two sparsity rates. For 100-token sequences, latency is reduced by 4.91$\%$ on the NVIDIA A100 and 5.81$\%$ on the NVIDIA Jetson Nano relative to the fastest baseline, CFSP. Future work will focus on improving search scalability through search-space partitioning and extending evaluation to more edge platforms.

\section*{Acknowledgment}
This project has been supported by CPER CornelIA of the region Haut de France and ANR project Exa-Ma ANR-22-EXNU-0002.
\section*{Declaration of AI-Tool Usage}
The authors used ChatGPT to improve grammar, sentence structure, and manuscript concision. They reviewed and edited the resulting text and take full responsibility for the manuscript.\par
\begingroup
\footnotesize
\setlength{\emergencystretch}{1em}

\endgroup
%

\end{document}